\documentclass[letterpaper, 10 pt, conference]{ieeeconf}

\IEEEoverridecommandlockouts
\usepackage{cite}
\usepackage{amsmath,amssymb,amsfonts}
\usepackage{algorithmic}
\usepackage{graphicx}
\usepackage{textcomp}
\usepackage{tabularx}
\usepackage[table,xcdraw]{xcolor}
\usepackage{booktabs}
\usepackage{booktabs}
\usepackage{colortbl}
\usepackage{multirow}
\usepackage{graphicx}
\usepackage[utf8]{inputenc}
\usepackage{authblk}
\usepackage{url}
\usepackage{float}
\usepackage{hyperref}
\usepackage{cleveref}
\usepackage[normalem]{ulem}
\newcommand{\modelicon}[2]{%
  \raisebox{-0.4ex}{\includegraphics[height=1.2em]{#1}}~#2%
}

\newcommand{\openai}[1]{%
  \raisebox{-0.4ex}{\includegraphics[height=1.2em]{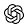}}~#1%
}

\newcommand{\deepseek}[1]{%
  \raisebox{-0.4ex}{\includegraphics[height=1.2em]{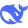}}~#1%
}

\newcommand{\gemini}[1]{%
  \raisebox{-0.4ex}{\includegraphics[height=1.2em]{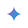}}~#1%
}

\newcommand{\defog}[1]{%
  \raisebox{-0.4ex}{\includegraphics[height=1.2em]{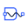}}~#1%
}

\newcommand{\claude}[1]{%
  \raisebox{-0.4ex}{\includegraphics[height=1.2em]{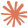}}~#1%
}

\newcommand{\qwen}[1]{%
  \raisebox{-0.4ex}{\includegraphics[height=1.2em]{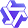}}~#1%
}

\newcommand{\ship}[1]{%
  \raisebox{-0.4ex}{\includegraphics[height=1.2em]{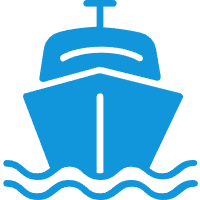}}~#1%
}

\newcommand{\duck}[1]{%
  \raisebox{-0.4ex}{\includegraphics[height=1.2em]{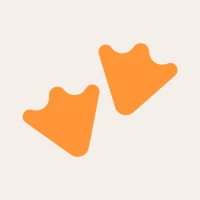}}~#1%
}

\newcommand{\ali}[1]{%
  \raisebox{-0.4ex}{\includegraphics[height=1.2em]{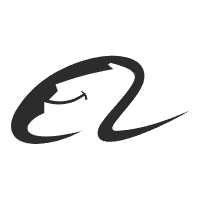}}~#1%
}

\newcommand{\llama}[1]{%
  \raisebox{-0.4ex}{\includegraphics[height=1.2em]{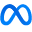}}~#1%
}

\def\BibTeX{{\rm B\kern-.05em{\sc i\kern-.025em b}\kern-.08em
    T\kern-.1667em\lower.7ex\hbox{E}\kern-.125emX}}
\title{\LARGE \bf VTS-LLM: Domain-Adaptive LLM Agent for Enhancing Awareness in Vessel Traffic Services through Natural Language}

\author{ \textbf{Sijin Sun}$^{1,2}$ \quad \textbf{Liangbin Zhao}$^{1,*}$ \quad \textbf{Ming Deng}$^{3}$ \quad \textbf{Xiuju Fu}$^{1}$
}
\affil{\normalsize 
$^*$Corresponding author \\
$^1$ Institute of High Performance Computing, Agency for Science, Technology and Research (A*STAR IHPC), Singapore \\
$^2$ National University of Singapore, Singapore \quad 
$^3$ Shanghai University, China
}

\begin{document}

\maketitle
\thispagestyle{empty}
\pagestyle{empty}

\begin{abstract}
Vessel Traffic Services (VTS) are essential for maritime safety and regulatory compliance through real-time traffic management. However, with increasing traffic complexity and the prevalence of heterogeneous, multimodal data, existing VTS systems face limitations in spatiotemporal reasoning and intuitive human interaction. 
In this work, we propose VTS-LLM Agent, the first domain-adaptive large language model (LLM) agent tailored for interactive decision support in VTS operations. We formalize risk-prone vessel identification as a knowledge-augmented Text-to-SQL task, combining structured vessel databases with external maritime knowledge. To support this, we construct a curated benchmark dataset consisting of a custom schema, domain-specific corpus, and a query–SQL test set in multiple linguistic styles.
Our framework incorporates NER-based relational reasoning, agent-based domain knowledge injection, semantic algebra intermediate representation, and query rethink mechanisms to enhance domain grounding and context-aware understanding. Experimental results show that VTS-LLM outperforms both general-purpose and SQL-focused baselines under command-style, operational-style, and formal natural language queries, respectively. Moreover, our analysis provides the first empirical evidence that linguistic style variation introduces systematic performance challenges in Text-to-SQL modeling. This work lays the foundation for natural language interfaces in vessel traffic services and opens new opportunities for proactive, LLM-driven maritime real-time traffic management.

\end{abstract}


\section{Introduction}

Vessel Traffic Services (VTS) serve as critical infrastructure deployed in major ports, straits, and coastal regions to ensure maritime safety, environmental protection, and the efficient management of vessel movements. These services rely on shipborne sensor data, such as AIS (Automatic Identification System) transmitted over VHF channels, along with shore-based radar observations and voice communication between operators and vessels, to provide real-time surveillance, navigational assistance, and regulatory enforcement.

\begin{figure}
    \centering
    \includegraphics[width=1\linewidth]{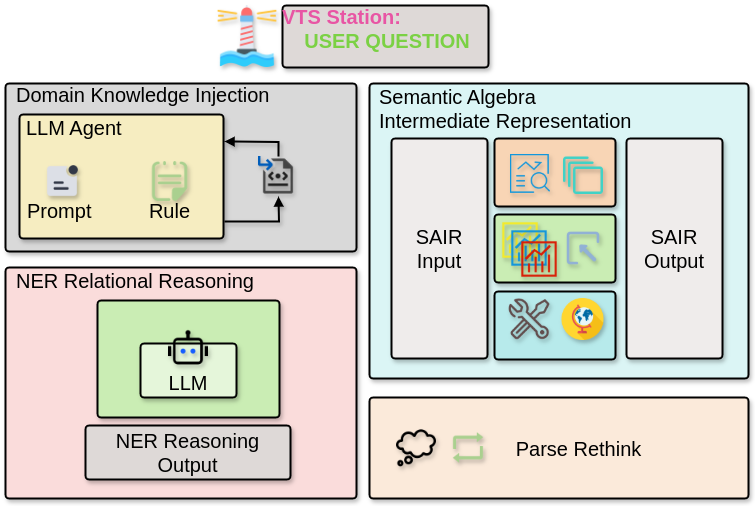}
    \caption{Simplified overall structure of the VTS-LLM agent.}
    \label{fig:overall_structure_simple}
\end{figure}
As global shipping volumes grow, VTS operations have become increasingly complex. Operators must continuously interpret dynamic, heterogeneous data—such as vessel movements, ship characteristics, voyage schedules, and geographic context—requiring ongoing spatiotemporal reasoning. This imposes a substantial cognitive load, particularly under time-critical conditions demanding rapid decisions and compliance assessments.

Although modern VTS systems support multi-source data fusion and visualized human-machine interaction, they remain limited in performing deeper spatiotemporal analysis. Tasks like early identification of risk-prone vessels—especially those involving abstract, localized rules—require integration of diverse databases and complex geospatial reasoning. Yet, current systems lack intelligent and intuitive mechanisms for exploratory querying and automated analysis. Vessel information retrieval still heavily depends on manual operator input and often requires formal SQL queries or navigating rigid interfaces, posing usability challenges and reducing efficiency.

Recent advances in the development and application of LLMs have shown promising potential to enhance various aspects of maritime transportation \cite{miller2025leveraging}, such as satellite image-based ship detection\cite{zhang2024popeye}, intelligent navigation\cite{wang2024kunpeng}, and the advancement of unmanned vessel technologies\cite{li2024robust}. While as for the aspect of real-time traffic management, the application of LLMs has been more extensively explored in land transportation domains\cite{zhang2024advancing}, such as traffic control \cite{li2024research,masri2025large} and spatiotemporal prediction\cite{yuan2024unist,wang2023building}. These advances illustrate the growing potential of LLMs to transform traffic management practices through data-driven, context-aware approaches.

Inspired by the work \cite{wang2024traffic}, which leverages LLMs for real-time land traffic surveillance through intelligent querying and contextual reasoning, we believe that, to the best of our knowledge, LLM-based agents now enable intuitive natural language exploration and context-aware analysis of maritime traffic data—capabilities not yet explored in VTS scenarios. To address the high cognitive load, we propose the first adaptation of an LLM-powered interactive agent VTS-LLM for awareness enhancement in VTS domain. 

Furthermore, motivated by geospatial demands of VTS operations and complexity of localized navigational rules, we reformulate risk-prone vessel identification as a knowledge-augmented Text-to-SQL task. Natural language queries on vessel behavior and rule compliance are grounded in both structured databases—capturing operational and spatiotemporal attributes—and external corpora containing domain knowledge and dynamic regulations. Based on this formulation, we construct a curated benchmark dataset to support future research in maritime-aware knowledge-augmented Text-to-SQL modeling in VTS scenarios.

Given domain-specific and knowledge-augmented characteristics of our application, effective Text-to-SQL performance hinges on the agent’s ability to generalize across specialized schemas, varied query intents, and heterogeneous contextual knowledge\cite{hong2024next}. Prior research on building more effective and adaptive Text-to-SQL LLMs has mostly focus on 1) in-context learning, including decomposition\cite{wang2025mac}, prompt optimization\cite{lee2025mcs} and chain-of-thought \cite{li2023can}; and 2) fine-tuning paradigm, including enhanced architecture\cite{kou2024cllms}, pretraining\cite{li2024codes}, data augmentation\cite{gao2023text} and multi-task tuning\cite{lee2025mcs}. These methods demonstrate excellent performance on the general-purpose evaluation benchmarks Spider1.0 \cite{yu2018spider1}, Spider2.0 \cite{lei2025spider2} and BIRD\cite{li2023can}.


However, prior studies have not addressed the unique knowledge-augmented characteristics and operational demands in maritime VTS scenarios. Beyond the challenges of spatiotemporal reasoning and knowledge integration, we identify key obstacles limiting LLM generalization. First, VTS operator expressions often follow compact, task-oriented styles with incomplete or non-standard syntax, differing from the well-formed language on which most models are trained. Second, vessel databases evolve over time and include ambiguous or overlapping entities (e.g., ship names, area codes), complicating entity linking and query interpretation.

To address these challenges in the VTS domain, we develop VTS-LLM with targeted adaptations: NER-based relational reasoning, agent-based domain knowledge injection, semantic algebra intermediate representation, and query rethink mechanisms. These components enhance schema alignment, semantic grounding, and rule interpretation, establishing a performance baseline for future benchmarking. We also propose a noval penalty-based evaluation metric that adjusts scores based on error severity, improving assessment in safety-critical scenarios. Experiments on our \href{https://huggingface.co/datasets/PassbyGrocer/vts-sql}{VTS-SQL} dataset show that VTS-LLM achieves 72.60\%, 77.80\%, and 89.72\% under command, operational, and formal natural language styles, respectively, demonstrating strong robustness across linguistic variations. Besides, to our best knowledge, our work presents the first empirical evidence that linguistic style variation can introduce significant and systematic challenges in Text-to-SQL modeling, an issue previously underexplored in the literature.


As summarized, our contributions are as follows:

1) Formalize a knowledge-augmented Text-to-SQL task for identifying risk-prone vessels in VTS system, by constructing a multimodal dataset for benchmark consisting of: (i) a custom-designed relational database capturing vessel traffic features, (ii) an external maritime knowledge corpus for VTS, and (iii) a query–SQL pair test-set covering different linguistic styles, including command-style, operational-style, and formal natural language style.

2) Propose VTS-LLM, the first LLM-based agent system adapted to interactive decision support in VTS operations, featuring a domain-adaptive framework with NER-based reasoning, semantic algebra intermediate representation, agent-based domain knowledge injection, and query rethink mechanisms. The agent model enables real-time, context-aware operational dialogue and achieves superior performance over general-purpose and SQL-specialized models.

3) Provide the first empirical evidence that linguistic style variation introduces systematic performance challenges in Text-to-SQL tasks, with LLM performing worse on concise, informal, and fragmented queries, highlighting an underexplored aspect in prior research.


\section{Vessel-of-Interest Identification in VTS via Text-to-SQL}
\subsection{Identification of Vessels of Interest}

VTS operations primarily involve responding to vessel inquiries and monitoring ships that may pose navigational risks. While basic information retrieval, such as checking vessel data via identifiers, is straightforward, the task of monitoring and identifying vessels that may pose navigational risks is significantly more complex. Operators must reason over large volumes of dynamic traffic data and detect subtle early-stage indicators of potential hazards. 

To address this, we reformulate the task as a Text-to-SQL problem, a well-established paradigm in the fields of artificial intelligence, enabling the translation of natural language queries from operators into executable SQL queries over structured vessel traffic database.

Unlike existing CPA-based (Closest Point of Approach) alert mechanisms, which highlight vessels only when collision risk is imminent, our focus is on earlier identification of vessels of interest—such as those approaching congested areas or experiencing prolonged signal loss—allowing for more proactive supervision.

Besides, VTS operations must also enforce localized navigational rules that restrict certain vessel types within specific maritime zones—particularly in high-traffic or constrained areas where factors like size, draft, or navigation behaviors affect traffic. For instance, very large crude carriers (VLCCs) and deep-draft vessels may be barred from entering certain channels during specific time windows. While not always posing immediate collision risks, violations of such rules can disrupt traffic order or compromise safety. Identifying these vessels is thus a key part of proactive supervision and falls within our broader definition of vessels of interest.

\subsection{Knowledge-augmented Text-to-SQL with Operational Linguistic Queries}

To better reflect practical VTS scenarios, we refine the problem into a knowledge-augmented variant of the Text-to-SQL task, tailored to the operational query styles used in VTS.

First, maritime supervision involves rich domain knowledge that extends beyond traffic data schema. This includes specialized maritime terminology, localized navigational rules and procedures specific to particular sea areas, as well as time-sensitive documents such as Notices to Mariners and real-time traffic advisories. These materials are essential for correctly interpreting operator queries and ensuring accurate rule-based supervision. Given their dynamic and context-dependent nature, such knowledge must be incorporated through knowledge-augmented mechanisms that can provide relevant textual guidance during query understanding and reasoning.

Second, operator-issued queries in VTS scenarios are typically short, fragmented, and task-oriented, deviating from standard natural language patterns. These utterances often omit syntactic elements, contain domain-specific shorthand, or follow operational communication protocols rather than grammatically complete sentences. As a result, models must demonstrate robustness to varied, informal, and context-dependent language expressions in order to correctly interpret intent and generate accurate SQL representations.

\subsection{Dataset}
To support development and evaluation, we construct a domain-specific multimodal dataset comprising: (1) a relational database, implemented by MySQL8.0 of structured vessel traffic data, as shown in \Cref{tab:data_table}; (2) a retrieval corpus of related maritime textual knowledge; and (3) A query–SQL pair test set containing diverse linguistic expressions and their corresponding executable SQL statements. These components reflect practical VTS operations and support realistic benchmarking of knowledge-augmented Text-to-SQL models. Dataset is available at \href{https://huggingface.co/datasets/PassbyGrocer/vts-sql}{VTS-SQL}.

\begin{table}[!htbp]
\centering
\caption{Simplified Description of Database Tables}
\begin{tabularx}{\linewidth}{llX}
\toprule
Table              & Columns & Description \\ \midrule
ship\_ais          & 17      & Latest AIS data of ships, including ship attributes (length, type, tonnage, etc.) and motion attributes (processed location, speed, heading, etc.). \\ 
ship\_ais\_quarter & 17      & AIS data sampled every 15 minutes with 10-second intervals, containing ship attributes and motion attributes. \\ 
shp\_data          & 6       & Geographic shape data (points \& polygons) stored in MySQL, including object type, name, and geometry. \\ 
warn\_single       & 11      & Records of ship encounter warnings. Includes CPA values and timestamps for each ship pair. \\ 
\bottomrule
\end{tabularx}
\label{tab:data_table}
\end{table}

The query set is designed based on input from professional VTS operators to reflect real-world operational scenarios, covering multiple query types, as demonstrated in \Cref{tab:question_set}. And, in addition to the operational style, each query is further provided in two additional linguistic variants: a concise command-style format and a more formal natural-language expression, as demonstrated in \Cref{tab:question_style}.


\begin{table}[!htbp]
\centering
\setlength{\tabcolsep}{6pt}  
\renewcommand{\arraystretch}{1.5}  
\caption{Query Types and Examples}
\begin{tabular}{>{\raggedright\arraybackslash}p{0.35\columnwidth} >{\raggedright\arraybackslash}p{0.55\columnwidth}}
\toprule
\textbf{Query Type} & \textbf{Example} \\
\midrule
Basic information retrieval & What is the current speed and location of vessel sounds like \textit{ALABAMA}? \\
Involving spatiotemporal reasoning & show me the ships in the waterway that may enter the port in the next an hour \\
Involving domain-specific terminology from the knowledge base & List MMSI and name of VLCCs and DDVs in the Strait. \\
Involving navigational rules from the knowledge base & list mmsi and names of all the vessels which against the speed requirements \\
Involving named entity recognition (NER) & where is \textit{West Coast}?\\
\bottomrule
\end{tabular}
\label{tab:question_set}
\end{table}

\begin{table}[htbp]
\centering
\small
\setlength{\tabcolsep}{6pt}
\renewcommand{\arraystretch}{1.3}
\caption{Linguistic Variant of the Query}
\resizebox{\columnwidth}{!}{
\begin{tabular}{>{\raggedright\arraybackslash}p{0.35\columnwidth} >{\raggedright\arraybackslash}p{0.6\columnwidth}}
\toprule
\textbf{Linguistic Style} & \textbf{Example} \\
\midrule
Operational & List MMSI and names of ships arriving port in next 30 min. \\ Command  & Arriving vessel in next 30 min? – list MMSI, name \\
Formal Natural Language & Could you show me the MMSI numbers and vessel names of ships expected to arrive at port within the next 30 minutes? \\
\bottomrule
\end{tabular}
}
\label{tab:question_style}
\end{table}

\section{Methodology}

Considering the complex spatial-temporal characteristics and the high dependence on expert domain knowledge inherent in maritime data analysis, we introduce a VTS domain-adaptive LLM (VTS-LLM), which incorporates four key methodological innovations to facilitate and optimize natural language to SQL conversion tasks for identifying vessels of interest in VTS scenarios. \Cref{fig:overall_structure_simple} illustrates the simplified modular architecture.


\subsection{NER-based Relational Reasoning}
 As the complexity and diversity in maritime terminology, we introduce an interactive Named Entity Recognition-based (NER-based) Relational Reasoning module aimed at clarifying and confirming relational semantics between the natural-language query entities and available database entities through iterative interactions.

\begin{figure}[!htbp]
    \centering
    \includegraphics[width=1\linewidth]{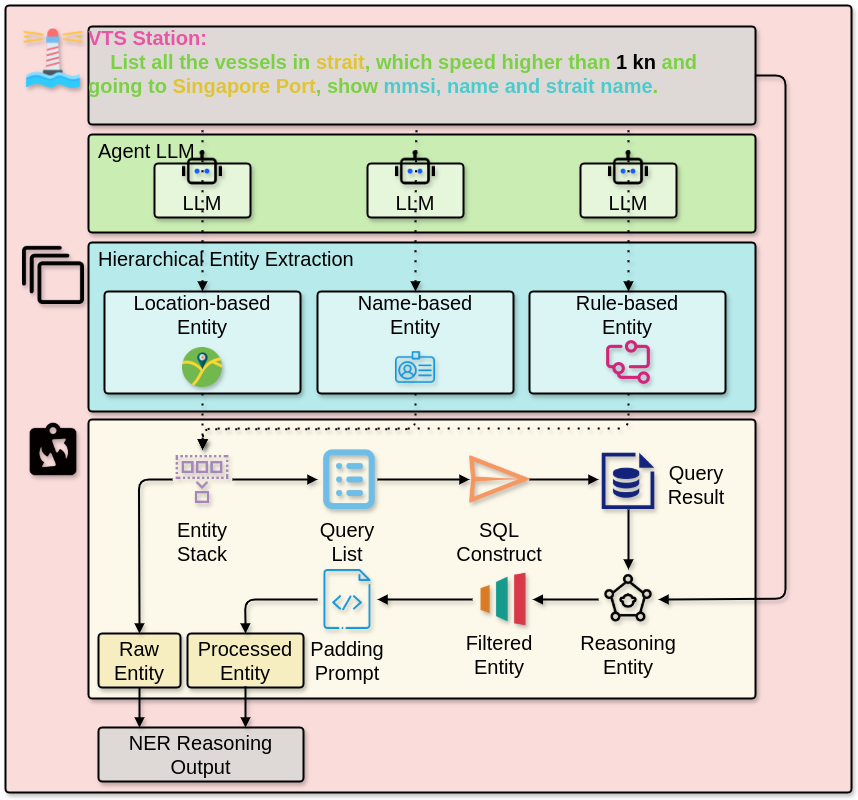}
    \caption{Hierarchical NER and Reasoning Framework for Maritime Query Processing}
    \label{fig:ner}
\end{figure}

The proposed module, shown in \Cref{fig:ner}, employs a hierarchical NER method tailored to maritime contexts, identifying and annotating entities in multiple semantic granularities. Broad categories include geographical regions, named types, and navigational facilities. Such hierarchical annotation provides structured representations for entities, significantly mitigating ambiguity compared to traditional flat NER schemes.

Recognized entities are subsequently utilized to build corresponding verification queries directed toward underlying spatial-temporal vessel traffic database. The query returns are then synthesized into structured, informative prompts, enabling LLMs to leverage CoT reasoning in determining and clarifying potential links between user-query entities and actual entities existing in database. For example, upon identification of entities such as \texttt{strait}, \texttt{Singapore Port} and \texttt{Pilot Station} verification queries are dynamically generated and executed to check the presence, relevance, and relational context of these entities within the database.


\subsection{Agent-based Domain Knowledge Injection}
%

\begin{figure}[!htbp]
    \centering
    \includegraphics[width=1\linewidth]{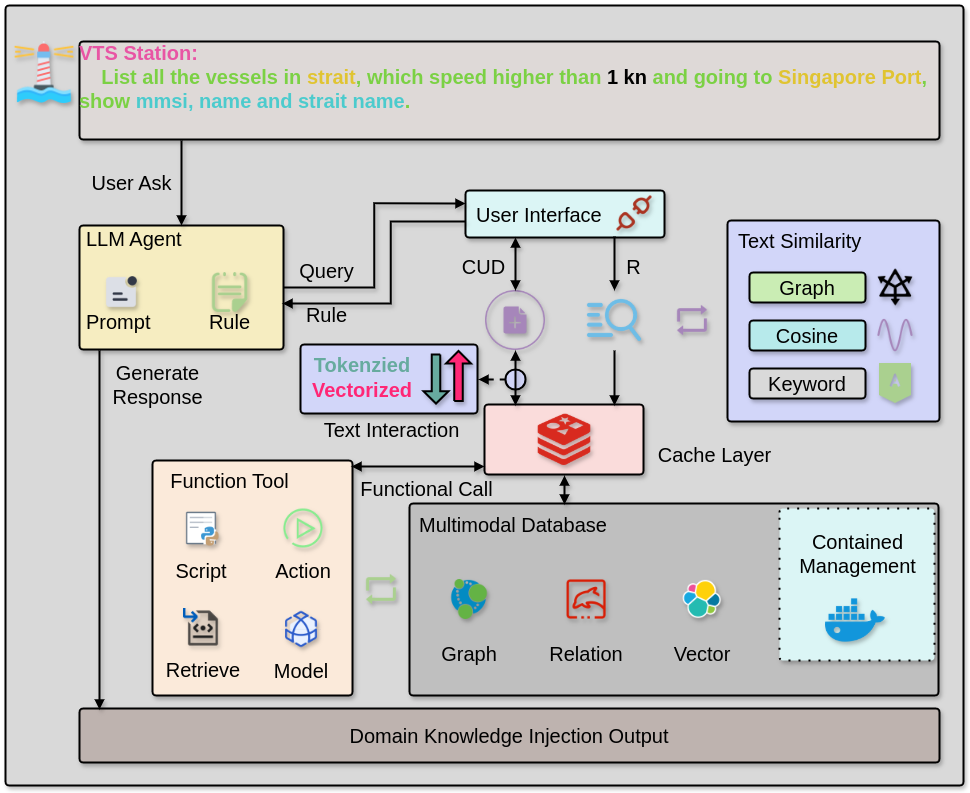}
    \caption{Agent-based Maritime Knowledge Injection Framework.}
    \label{fig:rule}
\end{figure}

The maritime domain involves extensive expert knowledge with complex semantic implications, including navigation planning, vessel type categorizations, geographic topology constraints, and port operation regulations. To effectively integrate such domain expertise into the Text-to-SQL process, we propose an agent-based framework to systematically operationalize maritime-specific knowledge. As illustrated in \Cref{fig:rule}, our framework dynamically introduces domain knowledge by intelligently decomposing a complex natural language query into a structured series of domain-specific subtasks, thus ensuring semantic clarity and operational effectiveness in the generated SQL queries.

\paragraph{Domain Knowledge-Aware Semantic Parsing} In the first stage, the system processes incoming natural language input with the support of Retrieval Augmented Generation (RAG)\cite{siriwardhana2023improving}. Using semantic matching, the agent retrieves relevant domain-specific information and converts it into structured semantic expressions with clear domain context. It then reformulates the natural language request into structured columns and values, followed by origin table structure.

\paragraph{Natural Language Processing and Interactive Query Engine} Following semantic parsing, this structured query information flows into an interactive query engine consisting of an LLM agent and user interface components. The LLM uses carefully-designed prompt strategies and internally-defined operational rules to further translate structured predicates into executable query instructions. Here, tokenization and vectorization facilitate standardized processing of the structured inputs. Efficient cache-based layer interactions enable rapid query responsiveness and user interactivity through Create, Update, and Delete (CUD) with retrieval (R) operations.

\paragraph{Multimodal DBs and Functional Call} Ultimately, these instructions activate functional call\cite{kim2024llm} interfaces that interact directly with a robust multimodal database, combining diverse storage modes such as graph, relational, and vector databases, all managed within a unified containerization framework. Complex functional tools like comprising scripts, actions, retrieval algorithms.

\subsection{Semantic Algebra Intermediate Representation}
Text-to-SQL tasks differ significantly from traditional database queries in the maritime spatiotemporal context. These differences mainly stem from the complexity of geospatial semantics, the ambiguity of temporal constraints, and the specialized terminology in the maritime field. To this end, we propose a Semantic Algebra Intermediate Representation (SAIR) method to bridge the semantic\cite{glenn2024blendsql} gap between natural language queries and structured SQL.

\begin{figure}[!htbp]
    \centering
    \includegraphics[width=1\linewidth]{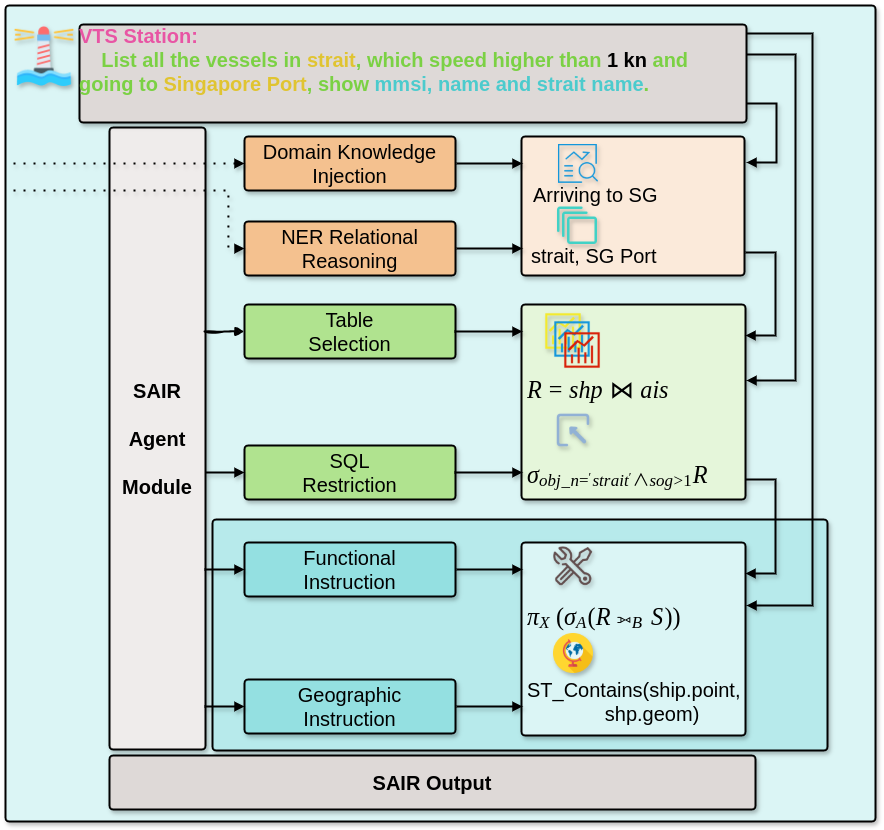}
    \caption{Schematic workflow of the SAIR module for VTS scenario.}
    \label{fig:sair}
\end{figure}

As shown in \Cref{fig:sair}, the SAIR module consists of three key submodules, which respectively handle named entity understanding and domain knowledge injection in orange module, structured semantic mapping in green module, and geographic function instructions in blue module. This design reflects the key improvement in our approach, which consists of decomposing the natural language parsing process into multiple substeps with clear structure and controllable semantics through explicit semantic modeling.

\paragraph{Semantic-aware Entity Understanding and Domain Enhancement} The input natural language query first undergoes NER reasoning, combined with maritime domain knowledge such as port names, waterway information, for semantic injection. This process forms semantic anchors by annotating the spatial location such as strait, and intent such as Arriving to SG, in the query, laying the foundation for subsequent algebraic representation.

\paragraph{Structured Algebraic Transformation Process} Customized relational algebraic expressions are introduced, including basic operations such as selection $\sigma$, projection $\pi$, and connection $\bowtie$, while integrating specific maritime semantic structures. 

\paragraph{Fusion of spatial and functional instructions} Relational algebra framework is extend and introduce spatial operations such as ST\_Contains and geographic matching instructions to express complex spatial semantics such as "whether the ship's position is in a specific waterway area". At the same time, the projection operation $\pi$ is used to implement attribute filtering to form the final output.

\subsection{Query Rethink}

Considering that the initially generated SQL queries may contain ambiguity, inaccuracy or grammatical errors (especially in complex spatiotemporal contexts), we propose the Query Rethink (RT) Mechanism, a novel validation and correction step\cite{wang2024rethinking} inspired by the thought chain paradigm. Specifically, given a SQL query draft, the model carefully checks its semantic coherence and logical consistency through an iterative reasoning process. By actively reconsidering the correctness of the generated SQL with domain-specific semantics and relational structures, the RT module identifies and dynamically corrects problematic query elements, ultimately generating refined and validated SQL statements.

\section{Experiment}

\begin{figure}[!htbp]
    \centering
    \includegraphics[width=1.0\linewidth]{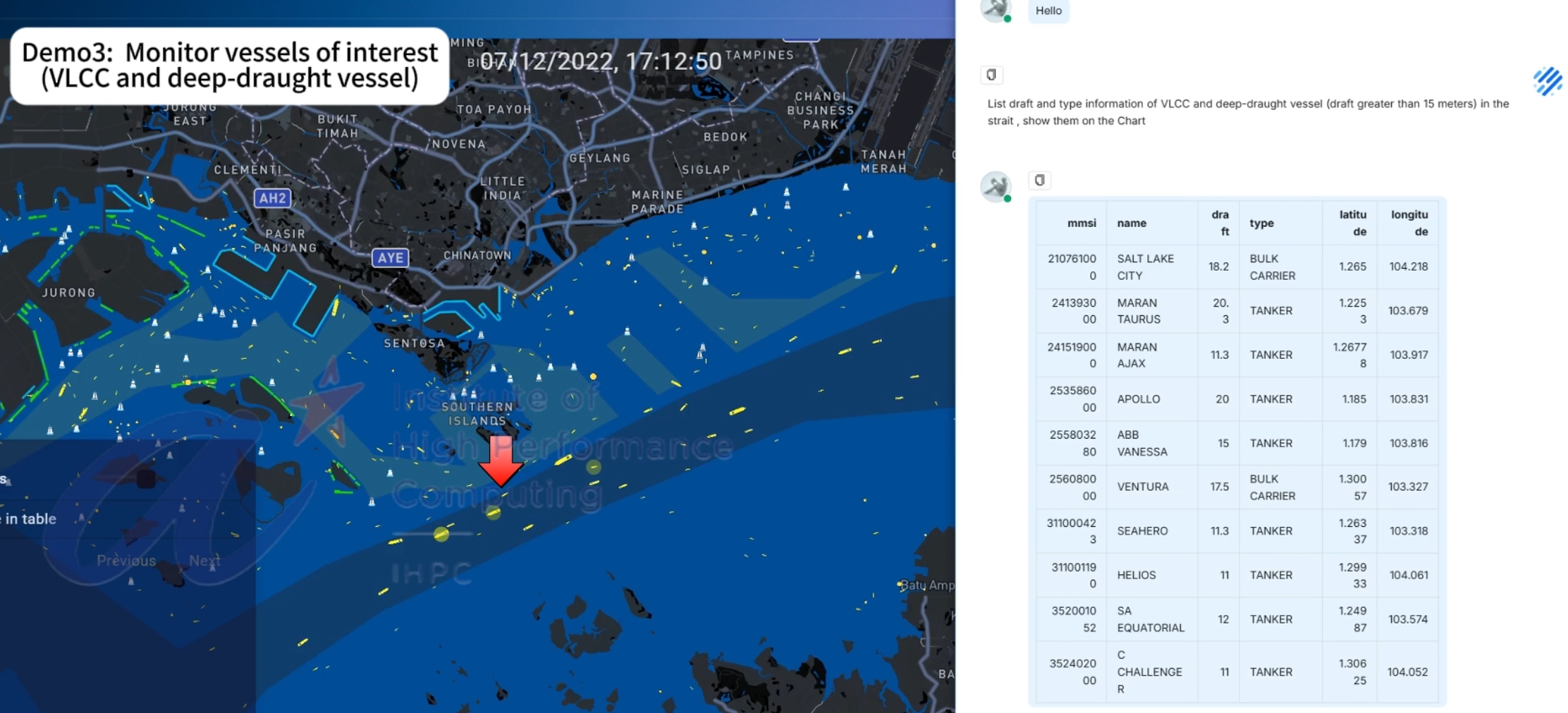}
    
    \vspace{0.1cm} 

    \includegraphics[width=1.0\linewidth]{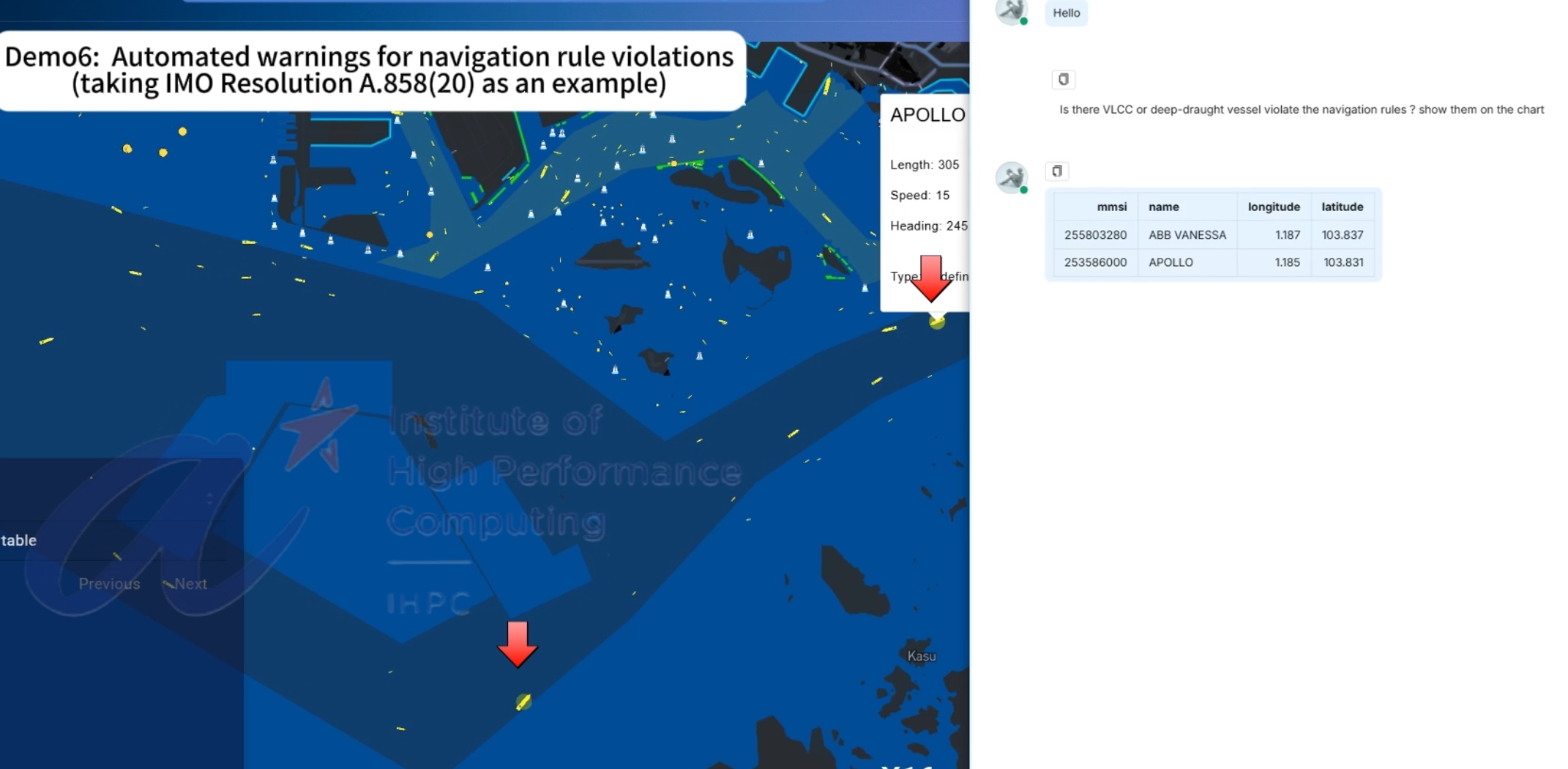}
    
    \caption{Demos of our developed system with VTS-LLM agent in VTS scenario.}
    \label{fig:demo_fig}
\end{figure}

\begin{table*}[!htbp]
\centering
\caption{Comparison Studies with Different Baseline LLMs. \textbf{Bold} and \uline{underlined} indicates the score ranks $1^{st}$ and $2^{nd}$.}
\begin{tabular}{@{}lcccccc@{}}
\toprule
                        & \multicolumn{5}{c}{Prompt Representation}                                    &                                        \\ \cmidrule(lr){2-6}
\multirow{-2}{*}{Model} & \modelicon{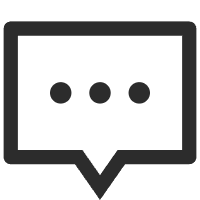}{Basic}          & \modelicon{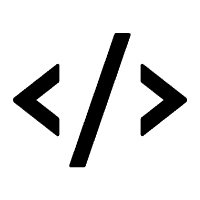}{Code}           & \modelicon{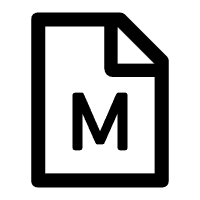}{Markdown}         & \modelicon{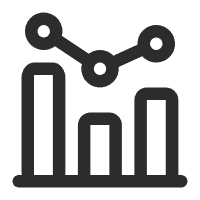}{Alpaca}         & \modelicon{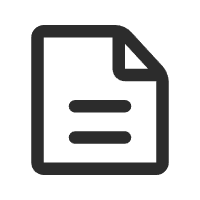}{Text}           & \multirow{-2}{*}{Avg.}                 \\ \midrule
\openai{GPT-4o-mini}             & 25.26          & 30.91          & 29.58          & 25.62          & 30.26          & 28.33                                  \\
\openai{GPT-4o}                  & 52.73          & 43.74          & 54.21          & 41.20          & 62.59          & 50.89                                  \\
\deepseek{DeepSeep-R1}             & 62.06          & \uline{64.02}    & 58.38          & 61.10          & 58.46          & 60.80                                  \\
\deepseek{DeepSeek-V3}             & 33.35          & 41.21          & 40.67          & 35.47          & 38.34          & 37.81                                  \\
\gemini{Gemini-2.0-Flash}       & 49.36          & 44.38          & 43.80          & 45.97          & 51.05          & 46.91                                  \\
\gemini{Gemini-1.5-Pro}          & 61.46          & 39.90          & 49.43          & 51.89          & 48.42          & 50.22                                  \\
\gemini{Gemini-1.5-Flash}        & 46.88          & 34.26          & 45.34          & 31.09          & 43.90          & 40.29                                  \\
\claude{Claude-3.7}              & \textbf{72.41} & 53.39          & \textbf{70.95} & \uline{64.39}          & \uline{65.03}          & \uline{65.23}                            \\
\qwen{Qwen-2.5}                & 45.35          & 42.87          & 43.36          & 50.72          & 48.94          & 46.25                                  \\
\qwen{Qwen-max}                & 49.01          & 49.97          & 59.99          & 45.53          & 52.30          & 51.36                                  \\ \midrule
                        & \uline{68.69}    & \textbf{68.76} & \uline{66.29}          & \textbf{74.30} & \textbf{66.53} & \textbf{68.91}                         \\
\multirow{-2}{*}{\ship{Ours}}  & \multicolumn{4}{c}{\openai{GPT-4o}}                                                         & Score &\cellcolor[HTML]{FFFF00}\textbf{77.80} \\ \bottomrule
\end{tabular}
\label{tab:online_llm}
\end{table*}

In experimental environment, we have completed the design and development of a working front-end system and back-end VTS
-LLM agent. Two demonstrations based on an embedded web platform have been developed to visualize maritime vessel information. As illustrated in \Cref{fig:demo_fig}, 
The demos showcase interactive querying by a VTS operator within a designated supervisory area, with vessel highlights automatically rendered on the front end by the backend LLM agent. Additional demonstrations can be viewed in the video available at \href{https://github.com/StanleySun233/maritime-vts}{vts-llm-demo}. The example queries used in the demos are as follows:
\begin{itemize}
    \item \texttt{List draft and type information of VLCC and deep-draught vessel in the strait, show them on the Chart}
    \item \texttt{Is there VLCC or deep-draught vessel violate the navigation rules? show them on the chart}
\end{itemize}

The remainder of this section presents the evaluation experiments of our methodology.

\subsection{Environment Setup}
The experiments were conducted on a high-performance computing system equipped with an Intel(R) Gold 6336Y CPU and NVIDIA RTX 4090 GPU with CUDA 12.1 support. The operating system was Ubuntu 22.04.

\subsection{Evaluate Metric}
Inspired by the \textbf{E}xact \textbf{M}atch (EM) and \textbf{EX}ecution Accuracy (EX) in SQL evaluation \cite{zhong2020semantic}, we aim to develop a customized evaluation metric specifically for SQL queries in the VTS domain. Existing metrics rely heavily on the exact match or execution correctness of result tuples, and fail to adequately address specific challenges and characteristics observed in vessel of interest identification, such as frequent redundant selections or missing relevant entries. Additionally, to balance precision and recall in safety-critical scenarios, we introduce a novel evaluation metric with penalty coefficients.

\Cref{eq:base_score} reflects the base score. Where $G^T$ represents the true result set, $G^P$ represents the predicted result set, and $|\cdot|$ represents the count of unique sample.

\begin{equation}
    B_s = \frac{|G^T \cap G^P|}{|G^T|}
    \label{eq:base_score}
\end{equation}

To further penalize predictions that introduce unnecessary over-selected tuples, we introduce a penalty factor $P_f$ as defined in \Cref{eq:penalty_factor}.

\begin{equation}
    P_f = \frac{1}{1 + \frac{|G^P| - |G^T|}{|G^T|}}, \quad |G^P|>|G^T|
    \label{eq:penalty_factor}
\end{equation}

Proposed Match Score $M_s$ in 
\Cref{eq:final_score} incorporates both matching and penalization conditions within an integrated metric.

\begin{equation}
    M_s =
\begin{cases}
B_s \cdot P_f, & |G^P| > |G^T| \\
B_s, & |G^P| \le |G^T| \\
0, & \text{otherwise}
\end{cases} \quad \cdot 100 \%
\label{eq:final_score}
\end{equation}

Designed $M_s$ effectively addresses distinct characteristics of this knowledge-augmented Text-to-SQL task by balancing recall on avoiding missing correct tuples and precision on discouraging unnecessary duplication.

\subsection{Comparison Studies}


As shown in \Cref{tab:online_llm}, we conduct the comparison experiment based on the queries of operational style in our dataset. For fairness, all compared models are given the same base knowledge and rules uniformly through prompt-based inputs. Details of prompt is available with code.

Results show that our VTS-LLM achieves an overall accuracy of 77.80\%, significantly outperforming SOTA baselines. And also it demonstrates superior performance across various prompt formats, particularly accounting for 74.30\% in Alpaca-style prompts, highlighting its strong generalisation and adaptability.

\begin{table}[!htbp]
\centering
\caption{Comparison of local-deployed \& Structured LLMs.}
\resizebox{\columnwidth}{!}{
\begin{tabular}{@{}lcccccc@{}}
\toprule
                        & \multicolumn{5}{c}{Prompt Representation}                                    &                                        \\ \cmidrule(lr){2-6}
\multirow{-2}{*}{Model} & \modelicon{./icon/basic.png}{Basic}          & \modelicon{./icon/code.png}{Code}           & \modelicon{./icon/markdown.png}{Markdown}         & \modelicon{./icon/alpaca.png}{Alpaca}         & \modelicon{./icon/text.png}{Text}           & \multirow{-2}{*}{Avg.}                 \\ \midrule
\multicolumn{1}{l|}{\deepseek{DS-Coder-v2:16b}} & 32.48          & 25.25          & 24.56          & 25.06          & \multicolumn{1}{c|}{26.35}          & 26.74          \\
\multicolumn{1}{l|}{\deepseek{DS-R1:7b}}        & 6.18           & 3.46           & 2.57           & 6.53           & \multicolumn{1}{c|}{4.87}           & 4.72           \\
\multicolumn{1}{l|}{\qwen{Qwen2.5-Coder:32b}}     & 41.34          & 48.97          & 58.26          & 47.79          & \multicolumn{1}{c|}{46.56}          & 48.58          \\
\multicolumn{1}{l|}{\qwen{Qwen2.5-Coder:7b}}      & 27.29          & 31.21          & 30.57          & 31.03          & \multicolumn{1}{c|}{18.94}          & 27.81          \\
\multicolumn{1}{l|}{\duck{DuckDB:7b}}             & 6.37           & 10.82          & 13.37          & 9.50           & \multicolumn{1}{c|}{21.34}          & 12.28          \\
\multicolumn{1}{l|}{\llama{LLama3.1:70b}}             & 42.65           & 47.78          & 48.81          & 27.80           & \multicolumn{1}{c|}{45.02}          & 42.41          \\
\multicolumn{1}{l|}{\llama{LLama3.3:70b}}             & 47.19           & 47.49          & 57.46          & 58.56           & \multicolumn{1}{c|}{52.90}          & 52.72          \\
\multicolumn{1}{l|}{\ship{Ours}}                  & \textbf{68.69} & \textbf{68.76} & \textbf{66.29} & \textbf{74.30} & \multicolumn{1}{c|}{\textbf{66.53}} & \textbf{68.91} \\ \midrule
\multicolumn{1}{l}{Structured}                                   & \multicolumn{4}{c}{Backbone LLMs}                                                                                &    &   Score         \\ \midrule
\multicolumn{1}{l|}{\ali{DAIL-SQL}}              & \multicolumn{5}{l|}{\openai{GPT-4o}
}                                                                             & 55.59          \\
\multicolumn{1}{l|}{\defog{SQLCoder:34b}}              & \multicolumn{5}{l|}{\llama{CodeLlama:34b}}                                                                         & 33.89          \\
\multicolumn{1}{l|}{\ship{Ours}}                  & \multicolumn{5}{l|}{\openai{GPT-4o}}                                                                             & \textbf{77.80} \\ \bottomrule
\end{tabular}
}
\label{tab:local_llm}
\end{table}

Besides, in \Cref{tab:local_llm}, our model maintains robust advantages over open-source and specialized structured-data LLMs, and surpassing two dedicated Text-to-SQL models, DAIL-SQL with GPT-4o and SQLCoder with LLama as backbone.

\subsection{Ablation Study}


To further analyze the effectiveness of each proposed component, we conducted ablation experiments as shown in \Cref{tab:ablation}. We separately drop three key modules in VTS-LLM: NER-based Relational Reaoning (NER), Semantic Algebra Intermediate Representation (SAIR), and Query Rethink mechanism (RT). 

\begin{table}[!htbp]
\centering
\caption{Ablation experiments of each components.}
\begin{tabular}{@{}ccccc@{}}
\toprule
No.  & NER & SAIR & RT & Score \\ \midrule
1    & $\surd$   & $\surd$  & $\surd$  & 77.80 \\
2    &    & $\surd$  &  $\surd$  & 72.45 \\
3    & $\surd$   &   &  $\surd$  &   74.06 \\
4    & $\surd$   & $\surd$  &    & 70.22 \\ \bottomrule
\end{tabular}
\label{tab:ablation}
\end{table}

The ablation study demonstrates the reduction of each core components in evaluation performance: the VTS-LLM agent (\#1) integrating all components achieves optimal performance (77.80\%); Without NER Reasoning modules (\#2), the performance degrades, with accuracy dropping to 74.25\%; Without SAIR modules (\#3) the performance degrades, with accuracy dropping to 74.06\%; Without RT modules (\#4) the performance degrades, with accuracy dropping to 70.22\%.
\subsection{Sensitive Analysis}

Furthermore, among all compared models, we select Claude, the best-performing model under the operational linguistic style, and conduct a sensitivity analysis across command-style and formal nature language style. Results in \Cref{tab:sensitive_analysis} reveal substantial variations in model performance across linguistic styles. A consistent pattern is observed across all models that performance degrades as query expressions become more concise and less formally structured.

\begin{table}[!htbp]
\centering
\caption{Sensitive Impact of Different Command Query Styles.}
\resizebox{\columnwidth}{!}{
\begin{tabular}{@{}lcccccc@{}}
\toprule
                        & \multicolumn{5}{c}{Prompt Representation}                                    &                                        \\ \cmidrule(lr){2-6}
\multirow{-2}{*}{Model} & \modelicon{./icon/basic.png}{Basic}          & \modelicon{./icon/code.png}{Code}           & \modelicon{./icon/markdown.png}{Markdown}         & \modelicon{./icon/alpaca.png}{Alpaca}         & \modelicon{./icon/text.png}{Text}           & \multirow{-2}{*}{Avg./Score}                 \\ \midrule
\multicolumn{7}{c}{\textbf{Operational}}                                                                                            \\ \midrule
\multicolumn{1}{l|}{\claude{Claude-3.5}}             & 66.48 & 45.41 & 69.92    & 61.74  & \multicolumn{1}{c|}{65.87} & 61.88                 \\
\multicolumn{1}{l|}{\claude{Claude-3.7}}             & 72.41 & 53.39 & 70.95    & 64.39  & \multicolumn{1}{c|}{65.03} & 65.23                 \\
\multicolumn{1}{l|}{\ship{Ours}}         &       &       &          &        & \multicolumn{1}{c|}{}      & \textbf{77.80}        \\ \midrule
\multicolumn{7}{c}{\textbf{Command}}                                                                                                    \\ \midrule
\multicolumn{1}{l|}{\claude{Claude-3.5}}             & 42.76 & 43.06 & 53.31    & 44.48  & \multicolumn{1}{c|}{54.06} & 47.53                 \\
\multicolumn{1}{l|}{\claude{Claude-3.7}}             & 52.78 & 39.52 & 54.21    & 46.63  & \multicolumn{1}{c|}{52.45} & 49.12                 \\
\multicolumn{1}{l|}{\ship{Ours}}         &       &       &          &        & \multicolumn{1}{c|}{}      & \textbf{72.60}        \\ \midrule
\multicolumn{7}{c}{\textbf{Formal}}                                                                                                \\ \midrule
\multicolumn{1}{l|}{\claude{Claude-3.5}}             & 64.48 & 71.33 & 75.02    & 79.43  & \multicolumn{1}{c|}{73.83} & 72.82                 \\
\multicolumn{1}{l|}{\claude{Claude-3.7}}             & 71.52 & 72.55 & 71.17    & 71.19  & \multicolumn{1}{c|}{77.19} & 72.72                 \\
\multicolumn{1}{l|}{\ship{Ours}}         &       &       &          &        & \multicolumn{1}{c|}{}      & \textbf{89.72}        \\ \bottomrule
\end{tabular}
}
\label{tab:sensitive_analysis}
\end{table}

In terms of model comparison, VTS-LLM consistently outperforms other baselines across various linguistic styles, demonstrating stronger robustness to language style variation. Particularly in the Command-Style setting, which is most representative of real-world VTS scenarios, VTS-LLM achieves a score of 72.40\%, significantly outperforming Claude-3.5 (47.53\%) and Claude-3.7 (49.12\%), both of which experience substantial performance drops in this challenging linguistic style. 

\subsection{Discussion}


1) VTS-LLM achieves SOTA performance in this knowledge-augmented Text-to-SQL task, outperforming general-purpose models by 12–35\% and exhibiting superiority over dedicated SQL models as well. While these models demonstrate strong performance on general datasets like BIRD and Spider, they remain inadequate for handling domain-specific queries and knowledge-augmented data contexts. The strength of VTS-LLM stems from a set of domain-specific adaptation strategies, including a novel semantic algebra for resolving column ambiguities, entity-relation reasoning for complex query understanding, agent-based domain knowledge fusion, and query rethink mechanisms for enhanced generation accuracy.

2) A notable performance disparity in the sensitive analysis indicate that different linguistic styles in our dataset pose systematically varying levels of difficulty for Text-to-SQL models, highlighting the need for robust linguistic adaptability. Moreover, our VTS-LLM demonstrates stronger robustness to linguistic style variations, making it better suited to the unstructured and often fragmented nature of queries that are representative of real-world scenarios. The lower scores of Claude models in this category suggest that they may struggle with the lack of formal structure in such prompts, highlighting the importance of specialized optimization for industrial use cases where clarity and brevity are prioritized over grammatical formality.

\section{Conclusion}
In this work, we developed VTS-LLM, the first domain-adaptive LLM-powered system tailored for interactive decision support in VTS operations. By formalizing risk-prone vessel identification as a knowledge-augmented Text-to-SQL task and constructing a curated, domain-specific dataset, we demonstrate that natural language interaction in operational linguistic style can effectively enhance situational awareness and support rule compliance monitoring in real-time vessel traffic services. In developing VTS-LLM , we tackled the domain-specific challenges in VTS scenarios by NER-based relational reasoning, semantic algebra intermediate representation, agent-based domain knowledge injection, and query rethink mechanisms. These structure propose knowledge integration strategies and domain-adaptive modeling techniques to enhance knowledge grounding, context-aware reasoning, and text generation capabilities. 

Evaluation experiment validate VTS-LLM's effectiveness and superiority in handling diverse linguistic queries and complex spatiotemporal reasoning, accounting for 72.60\% on command-style, 77.80\% on operational-style and 89.72\% on formal natural language style. By integrating structured vessel databases with external textual knowledge, VTS-LLM enables more robust and context-aware reasoning for data exploration—going beyond basic information retrieval in conventional VTS systems. Besides, our work presents the first empirical evidence that linguistic style variation can introduce significant and systematic challenges in Text-to-SQL modeling, an issue previously underexplored in the literature.

Future work will focus on extending VTS-LLM to support a broader range of VTS and maritime regulations, improving real-time robustness in dynamic environments, and exploring joint optimization strategies about RAGs for enhanced query understanding. In addition, improving the model’s understanding of concise and informal query expressions remains an important and application-relevant challenge, which we consider a promising direction for future development in VTS domain. Besides, more proactive VTS activities could be supported by an LLM-based agent. For instance, the agent could interpret vessel calls, generate automated voice responses, perform basic radio duties, maintain relevant communication logs, and reply to standard messages on behalf of the operator.



\end{document}